\documentclass[10pt,twocolumn,letterpaper]{article}

\pdfoutput=1

\usepackage{wacv}
\usepackage{times}
\usepackage{epsfig}
\usepackage{graphicx}
\usepackage{amsmath}
\usepackage{amssymb}

\wacvfinalcopy 

\def\wacvPaperID{5} 

\ifwacvfinal\fi
\begin{document}

\title{Detecting GAN-generated Imagery using Color Cues}

\author{Scott McCloskey and Michael Albright \\
Honeywell ACST\\
{\tt\small \{scott.mccloskey, michael.albright\}@honeywell.com}
}

\maketitle
\ifwacvfinal\thispagestyle{empty}\fi

\begin{abstract}
Image forensics is an increasingly relevant problem, as it can potentially address online disinformation campaigns and mitigate problematic aspects of social media.  Of particular interest, given its recent successes, is the detection of imagery produced by Generative Adversarial Networks (GANs), e.g. `deepfakes'.  Leveraging large training sets and extensive computing resources, recent work has shown that GANs can be trained to generate synthetic imagery which is (in some ways) indistinguishable from real imagery.  We analyze the structure of the generating network of a popular GAN implementation \cite{nVidia}, and show that the network's treatment of color is markedly different from a real camera in two ways.  We further show that these two cues can be used to distinguish GAN-generated imagery from camera imagery, demonstrating effective discrimination between GAN imagery and real camera images used to train the GAN.
\end{abstract}

\section{Introduction}

With the increasing importance of social media as a means of disseminating news, online disinformation campaigns have gotten significant attention in recent years.  The dated phrase `seeing is believing' is still descriptive of how people validate such stories, though, so image forensics is increasingly important.  While social media makes the dissemination of fake news easier, computer vision tools have contributed to this trend by making it easier to generate fake imagery.  Whereas an image manipulator in prior years would need significant experience with rendering and/or image manipulation software, modern data-driven approaches have made it much easier to generate artificial imagery from scratch.

Our paper concerns the development of forensics to detect imagery from Generative Adversarial Networks (GANs).  While these are indistinguishable from real imagery {\em to the GAN's discriminator}, they differ in important ways from images taken by a camera.  We analyze the structure of the generator network, paying particular attention to how it forms {\em color}, and note two important differences.  First, the generator's internal values are normalized to constrain the outputs, in a way which limits the frequency of saturated pixels.  Second, the generator's multi-channel internal representation is collapsed to red, green, and blue channels in a way that's similar to models of color image formation, but uses weights that are quite different than the analogous spectral sensitivities of a camera.  We investigate the effectiveness of these two cues in detecting two types of GAN imagery: one being imagery wholly generated by a GAN and the other where GAN-generated faces replace real faces in a larger image.

\begin{figure}[t]
\begin{center}
\includegraphics[width=0.8\linewidth]{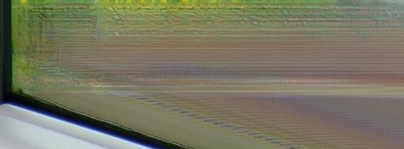}
\includegraphics[width=0.8\linewidth]{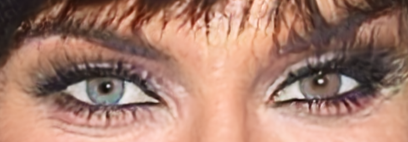}
\end{center}
\caption{Example artifacts evident in GAN-generated imagery.  Top image shows checkerboard artifacts introduced by deconvolution steps.  Bottom image shows mismatched eye colors, similar to a cue used in existing forensics \cite{albanyEyes}.}
\label{fig:artifacts}
\end{figure}

Our approach is based on an analysis of the GAN generator architecture, and is complementary to approaches that detect visual artifacts in the synthesized imagery.  While these two types of approaches are complementary, the rapid pace of GAN developments will likely mitigate the effectiveness of artifact-based detection.  In particular, the checkerboard artifacts illustrated in Fig. \ref{fig:artifacts} have already been mitigated \cite{checkerboard} by replacing deconvolution steps with up-sampling followed by convolution steps.  Additional cues, such as mismatched eye colors or a lack of blinking in GAN-generated video \cite{albanyEyes} are similarly likely to be eliminated. 

\section{Related Work}

Since their introduction in 2014 \cite{GAN_NIPS}, GANs have quickly become an extremely valuable tool in a range of computer vision applications.  At a high level, the concept of a GAN is that two networks are trained to compete with one another.  The `generator' network is trained to produce artificial imagery that is indistinguishable from a given dataset of real imagery, whereas the `discriminator' is trained to correctly classify imagery as being either real or coming from the generator. Early attempts at this \cite{NIPS2016_6125} were able to generate convincing imagery of simple image datasets such as MNIST digits \cite{MNIST}, but had a harder time mimicking more complicated images.  More recently, computational techniques have been introduced which can generate convincing facial imagery \cite{nVidia} and have increased the resolution of generated imagery \cite{wang2018pix2pixHD}.

In response to the development of GANs, the forensics community has begun to develop methods to detect whether or not a given image was generated by a network trained in a GAN framework (for brevity, we refer to the detection targets as `GAN images', and seek to distinguish them from `real images').  One such method \cite{albanyEyes} uses the lack of blinking in DeepFake-type videos to detect GAN videos.  Other approaches, rather than leveraging semantically-meaningful cues, use machine learning and neural networks to distinguish GAN from real images.  Marra \etal \cite{marra} use a network based on XceptionNet, Hsu \etal \cite{hsu} develop a deep forgery discriminator with a contrastive loss function, and Guera and Delp \cite{guera} use recurrent neural networks to detect GAN video.  A key concern with methods based on deep networks is that they could easily be incorporated into the GAN's discriminator and, with additional training, the generator could be fine-tuned in order to learn a counter-measure for any differentiable forensic.  An exclusively learning-based approach also makes it difficult to explain the outputs of the network, which is necessary in some forensics applications.  Our approach, which is complementary to the above-mentioned forensics, is to analyze the structure of the GAN's generator and see how it impacts image statistics.

\section{GAN Generator Architecture}
\label{sec:architecture}

In this section, we review the network architecture of the GAN's generator, and propose two specific cues which could be used to distinguish GAN imagery from real imagery.  Since generators use different structures from one GAN to another, we look at features which are common among GANs.  Also, in order to improve the detectability, we focus our attention on the later layers of the generator, since cues introduced in these layers are less likely to be modulated by subsequent processing.  

\begin{figure}[t]
\begin{center}
\includegraphics[width=0.9\linewidth]{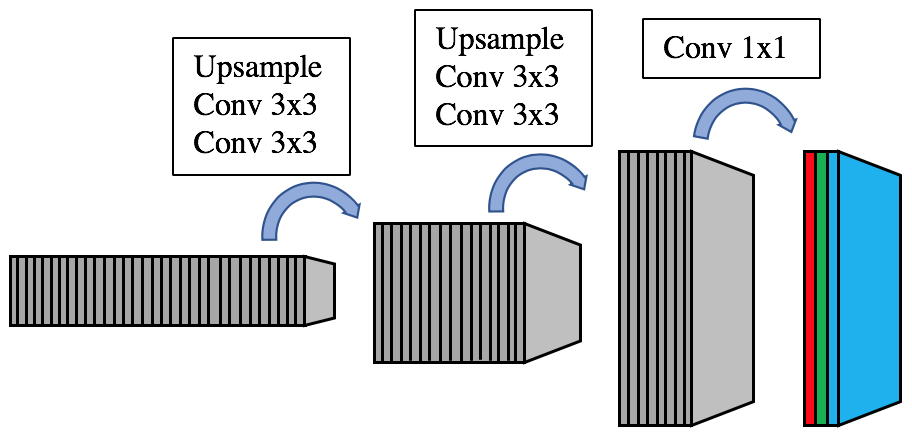}
\end{center}
\caption{Example of the generator architecture from \cite{nVidia}.  The high-resolution image is produced from an input `latent vector' by repeated upsampling (doubling the spatial dimensions), followed by 3x3 convolutions with leaky-ReLU activations and pixel-wise normalization. The final color image is generated by a 1x1 convolution.}
\label{fig:architecture}
\end{figure}

Figure \ref{fig:architecture} shows a representative generator architecture, which expands a learned feature representation of the desired class to an image of such an object.  The last layer of the generator, in particular, produces  
a 3-by-$W$-by-$H$ output array: an image having 3 color channels, $W$ columns, and $H$ rows.   The {\em input} to this last layer is an array of size $K$-by-$W$-by-$H$, where the $K>3$ layers are referred to as `depth' layers.  
The conversion of the $K$ depth layers to the red, green, and blue color channels provide two potential forensics cues, which will be described in the following sections.

\subsection{Color Image Formation}

\begin{figure*}[t]
\begin{center}
\includegraphics[width=0.3\linewidth]{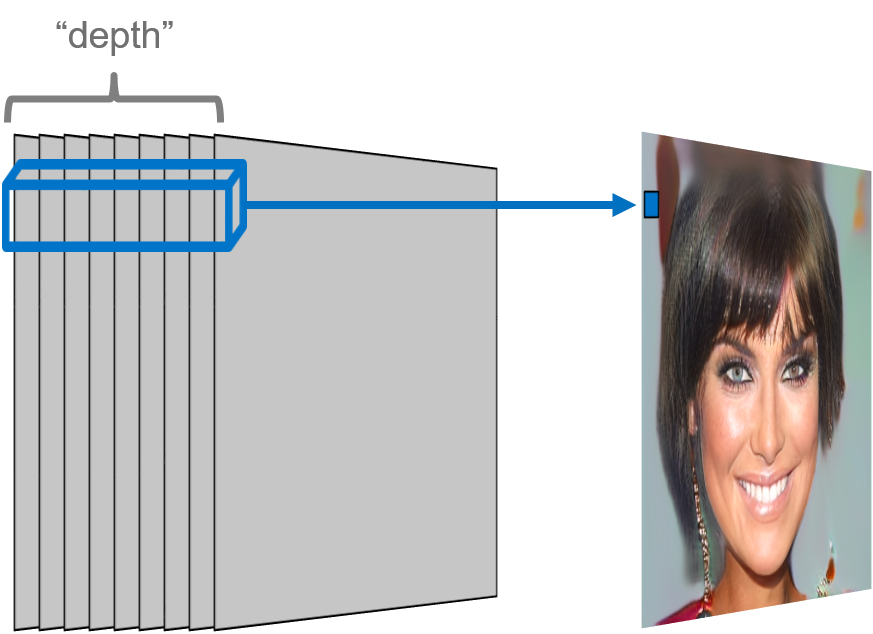}
\includegraphics[width=0.3\linewidth]{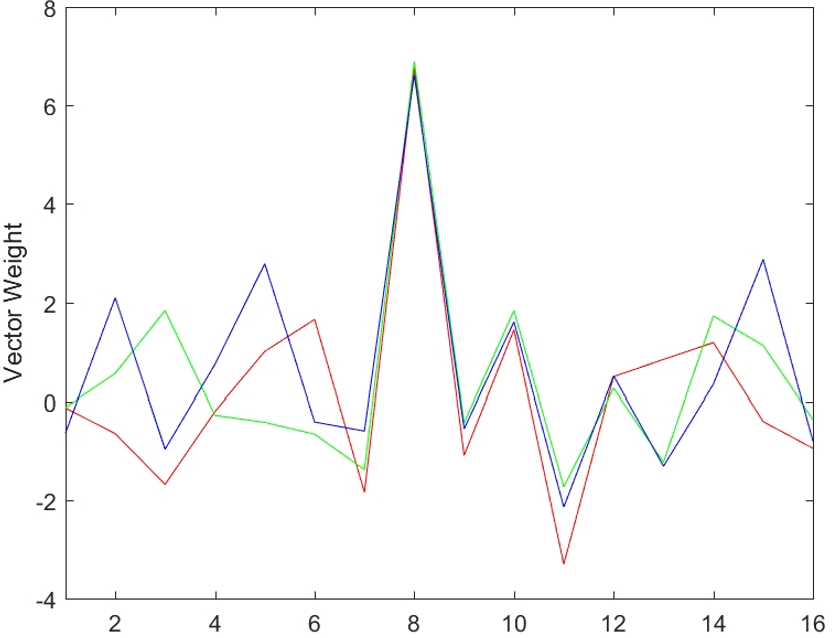}
\includegraphics[width=0.3\linewidth]{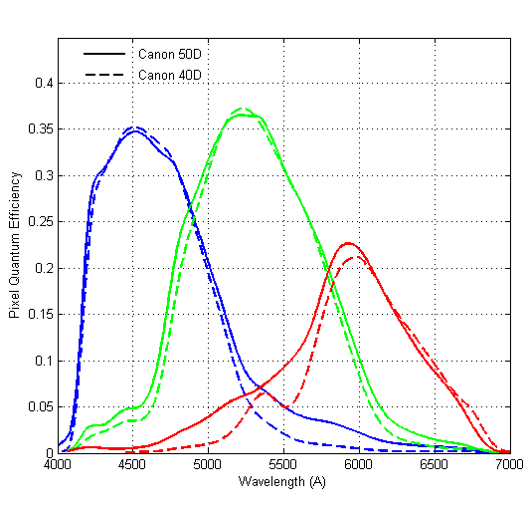}
\end{center}
\caption{(Left) The last layers of a GAN's generator collapse multiple `depth' layers to red, green, and blue pixel values via convolutions that span the depth layers, but have limited spatial extent.  (Center) The weights used for face image synthesis in \cite{nVidia} to collapse 16 depth layers to red, green, and blue are plotted.  (Right) By contrast, the spectral responses of real cameras' color filter arrays \cite{CanonSpectral} vary from camera to camera, but have a structure which is quite different than the learned weights of a GAN.}
\label{fig:colorFigure}
\end{figure*}

To the extent that the last network layer collapses $K>3$ depth layers to red, green, and blue color channels, it is similar to the mechanism by which a camera's color filter array integrates light over three wavebands to form a color image \cite{HuntBook}.  Figure \ref{fig:colorFigure} illustrates this analogy.  The multiple depth layers are combined in a weighted sum to create a color value at each pixel, with the weights being uniform over the spatial extent of the output.  In some cases, e.g. \cite{nVidia}, the last convolution has a non-singleton dimension in only the depth direction (Figure \ref{fig:colorFigure} left); in others, e.g. \cite{wang2018pix2pixHD}, the convolution has a larger spatial extent, as well. 

When the color filter array on a camera's sensor collapses the visible spectrum to RGB values, the `weighting' of light at different wavelengths is represented by the spectral response function.  Spectral response functions vary from camera to camera, but are driven by several constraints:
\begin{enumerate}
\item In order to allow for saturation and eliminate cross-talk, the spectral response functions for red, green, and blue have limited overlap.
\item Because the sensor counts photons that pass through the color filter array, spectral response functions must be non-negative.
\end{enumerate}
Neither of these constraints apply in the case of a synthetic generator, which doesn't need to count photons.  By allowing negative weights, saturation can result {\em even if there is significant overlap between the weights learned for different channels}.  Indeed, Fig. \ref{fig:colorFigure} (center) shows the 16 weights learned to synthesize face images in \cite{nVidia}, which share a common peak across the three color channels and are correlated at several non-peak values, as well.  By contrast, Fig. \ref{fig:colorFigure} (right) shows the spectral response functions for two different Canon cameras, which have different peak wavelengths for each channel and relatively less overlap in their sensitivities.

\subsection{Normalization}

\begin{figure*}[t]
\begin{center}
\includegraphics[width=.9\linewidth]{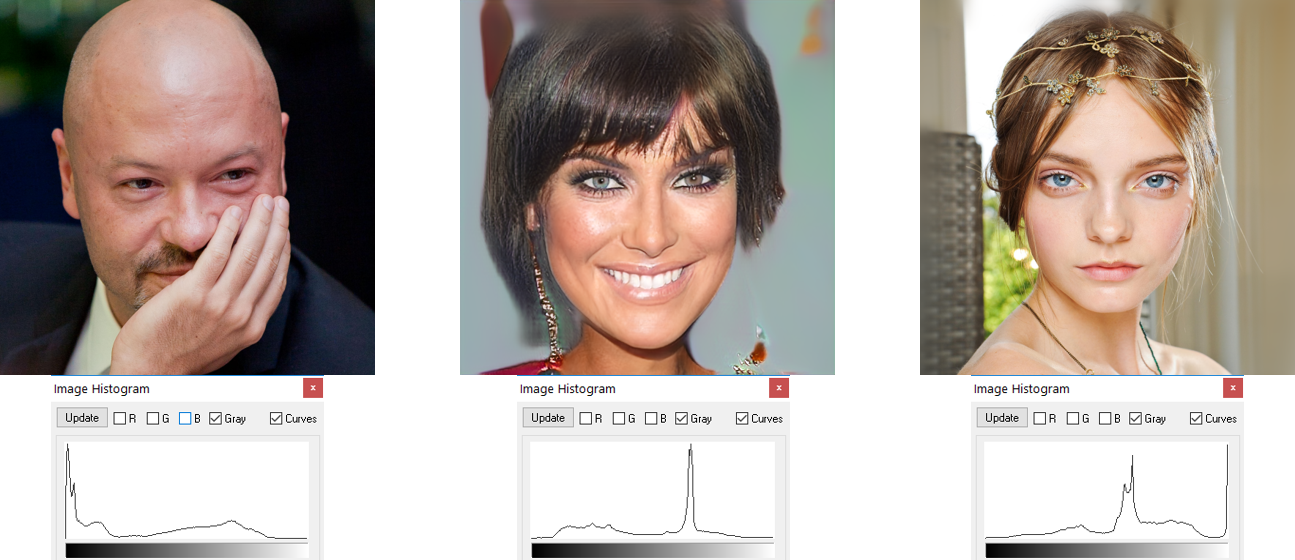}
\end{center}
\caption{Example images (top row) and grayscale histograms (bottom row) for two real images (left, right) and one GAN image (center) from \cite{nVidia}.  Whereas the real images feature regions of under- or over-exposure (left and right images, respectively), GAN images (e.g., center) lack regions of saturation even when the background is white.}
\label{fig:saturationFigure}
\end{figure*}

Another common process in GAN generators is the use of normalizations, which are employed in order to encourage convergence in training.  As with color image formation, the exact method of normalization varies from one GAN to the next.  In \cite{nVidia}, pixel-wise normalization is applied
%
after convolution layers, so that the values of the `depth' vector at each pixel have a fixed magnitude, i.e.
\begin{equation}
  b_{j,x,y} = \frac{ a_{j,x,y} }{ \sqrt{ \frac{1}{N} \sum_{c=0}^{N-1} (a_{c,x,y})^2 + \epsilon   } } \, ,
\end{equation}
where $a$ is an unnormalized feature map, $b$ is the pixel-wise normalized version, indices $x$ and $y$ denote the spatial location of the pixel, indices $j$ and $c$ denote the depth position in the feature map, $N$ denotes the number of feature maps, and  $\epsilon = 10^{-8}$.
In \cite{wang2018pix2pixHD}, normalization is applied within the individual `depth' planes as,
\begin{equation}
  b_{n,c,x,y} = \gamma_{n,c}\left( \frac{ a_{n,c,x,y} - \mu_{n,c} }{\sigma_{n,c}} \right) + \beta_{n,c} 
\end{equation}
\begin{equation}
  \mu_{n,c}  = \frac{1}{HW} \sum_{x=0}^{W-1}\sum_{y=0}^{H-1} a_{n,c,x,y}  
\end{equation}
\begin{equation}
  \sigma_{n,c} = \sqrt{ {1}/{HW} \sum_{x=0}^{W-1}\sum_{y=0}^{H-1} \left( a_{n,c,x,y} - \mu_{n,c} \right)^2 + \epsilon  } \, ,
\end{equation}
where $b$ and $a$ are the normalized and unnormalized feature maps (respectively), $x$ and $y$ specify the spatial position of the pixel, $c$ denotes the depth channel, $n$ indexes the image in the mini-batch, and the parameters $\beta$ and $\gamma$ are learned  during the training process to constrain the mean and variance of values within the feature map depth planes.

Regardless of whether the normalization is applied pixel- or layer-wise, the result of both steps will be to have a relatively uniform distribution in the unit interval.  These well-behaved values are then transformed into RGB intensities as described above.  In camera-based imaging, however intensity values are not nicely constrained.  Instead, irradiance values incident on a camera's sensor generally have a logarithmic distribution, necessitating high dynamic range (HDR) imaging \cite{HDRbook}.  HDR imaging involves the capture of multiple images separated by one or more {\em stops} (binary orders of magnitude) of exposure, e.g. images exposed for 1/15, 1/30, and 1/60 second.  Without HDR, camera images generally have regions of saturation and/or under-exposure, as shown in Fig. \ref{fig:saturationFigure}.  Because of the normalizations applied in the generator, however, GAN images lack these regions.

\section{Detection Methods}

In this section, we propose detection methods based on each of the above analyses, in order to understand the predictive power of these cues.  Given a relatively small set of training data, it is necessary to utilize pre-trained models (where applicable) or to use lower dimensionality features that can be trained with the data on hand.

\subsection{Color Image Forensics}

Intuitively, the overlap between the weights which map depth layers to RGB colors shown in Fig. \ref{fig:colorFigure} (center) should manifest themselves in a high correlation between color channels at a given pixel than would be evident from a real camera having spectral sensitivities similar to the Canon curves shown in Fig. \ref{fig:colorFigure} (right).  To investigate this, we use the standard rg chromaticity space, where 
\begin{equation}
r=\frac{R}{R+G+B}~~~~and~~~~g=\frac{G}{R+G+B}.
\end{equation}
We expect that GAN images will have higher than normal correlations in this space, but that the correlations will have no spatial component due to the fact that the color conversion (at least in the case of \cite{nVidia}) is applied independently at each pixel.  The evaluate the effectiveness of this cue for forensics, we adapt the method of Chen \etal \cite{CanChenCVPR2018}, which uses bivariate histograms for a forensics task.  In that work, the authors demonstrate that pixel-wise statistical relationships between intensity and noise can be used to detect focus manipulations, by building Intensity Noise Histograms (INH) which are classified by a deep network similar to 
VGG (\cite{VGG}).
In our case, the $r$ and $g$ chromaticity coordinates serve as the two variables, and the INH network is used classify these histograms as being a GAN image or a camera image.

We use a pre-trained version of INH from \cite{CanChenCVPR2018}, and fine-tune the classifier with $r$-vs.-$g$ histograms from a set of GAN imagery produced by \cite{nVidia} and a set of camera imagery used in the training of the GAN.

\subsection{Saturation-based Forensics}

For this forensic, the hypothesis is that the frequency of saturated and under-exposed pixels will be suppressed by the generator's normalization steps.  This suggests a straightforward GAN image detector, where we simply measure the frequency of saturated and under-exposed pixels in each image.  Specifically, for over-exposed pixels we measure a set of features
\begin{equation}
f_i^o = \frac{1}{HW} \| \{ (x, y)~|~I(x,y)\ge \tau_i^o \} \|,
\end{equation}
for $\tau_i^o \in \{240, 245, 250, 255\}$ (for an 8 bit representation of pixel intensities).  Similarly, the frequency of under-exposed pixels are measured as features
\begin{equation}
f_i^u = \frac{1}{HW} \| \{ (x, y)~|~I(x,y)\le \tau_i^u \} \|,
\end{equation}
for $\tau_i^u \in \{0, 5, 10, 15\}$.

These features are classified by a linear Support Vector Machine (SVM), trained using Matlab's fitcsvm function.  The training data consisted of features from 1387 GAN-generated images (randomly sub-sampled from the 
30 LSUN \cite{LSUN}  categories of images 
created by the GAN in \cite{nVidia}), and real camera images from the ImageNet dataset.

\section{Evaluation}

In order to evaluate the effectiveness of our two detection methods, we experimented with two benchmark datasets produced in conjunction with the US National Institute of Standards and Technology's Media Forensics Challenge 2018 \cite{MFC18}.  The two datasets address two different sets of GAN imagery:
\begin{enumerate}
\item {\bf GAN Crop} images represent smaller image regions which are either entirely GAN-generated or not.  
\item {\bf GAN Full} images are mostly camera images, but some faces have been replaced by a GAN-generated face, similar to deep fakes.
\end{enumerate}
For both datasets, we compute the features (histograms or saturation count features) over the entire image, even though GAN Full images have small manipulated regions around faces.  Following on the convention, we present our detectors' performance via a Receiver Operator Characteristic (ROC) curve, showing the true detection and false alarm rate as a function of a decision threshold applied to the continuously-varied score output by each classifier.  In an ROC curve, the performance of a random classifier would be a diagonal line.  We also summarize the ROC with its Area Under the Curve (AUC), which would be 0.5 for a random detector and 1 for a perfect detector.  

\subsection{Saturation Statistics}

\begin{figure}[t]
\begin{center}
\includegraphics[width=0.75\linewidth]{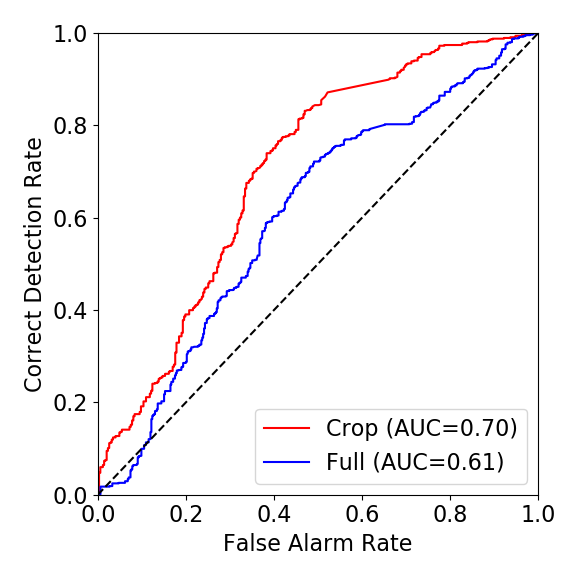}
\end{center}
\caption{ROC curves showing the performance of the saturation frequency SVM on the two GAN datasets from \cite{MFC18}.}
\label{fig:saturationROC}
\end{figure}

Figure \ref{fig:saturationROC} shows the ROC curve for our SVM trained on over-exposure features $f^o$.  For both datasets, the performance is significantly better than a random detector.  The method clearly does a better job of detecting fully GAN-generated images, where it produces a 0.7 AUC. In part this follows from a better match to the images used in training, but also the features' measuring a {\em proportion} of saturated pixels will be further diluted by the non-GAN regions in the GAN Full images.  Despite this, the method still produces a respectable ROC and 0.61 AUC on the GAN Full image set.

\begin{figure}[t]
\begin{center}
\includegraphics[width=0.75\linewidth]{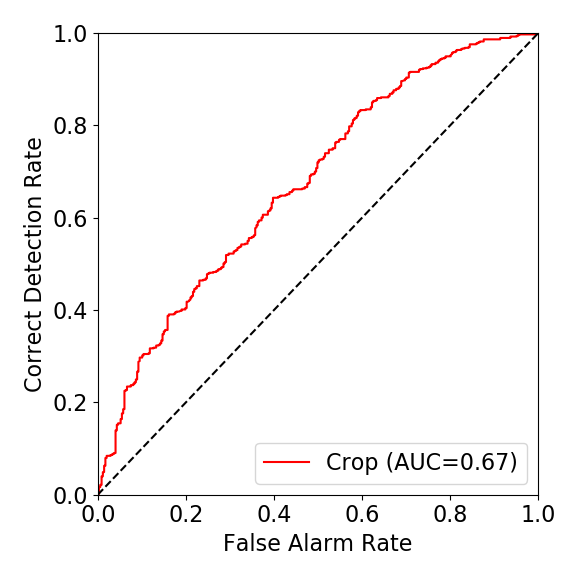}
\end{center}
\caption{ROC curves showing the performance of the saturation {\em and under-exposed} frequency SVM on GAN crop.}
\label{fig:saturationROC2}
\end{figure}

Interestingly, as shown in Fig. \ref{fig:saturationROC2}, the performance of the method decreases when the SVM is presented a feature vector consisting of {\em both} under- and over-exposed pixel frequencies, i.e. $f^u$ and $f^o$.  The AUC is reduced from 0.70 to 0.67, and has a notably lower correct detection rate at a 0.5 false alarm rate.  Though it's not immediately obvious why the under-exposed features should be less predictive than over-exposed pixel frequencies, one hypothesis is the asymmetric shape of the Rectified Linear Units (ReLU) as activations in \cite{nVidia}.  That said, the SVM could have learned to ignore the additional features from $f^u$, but failed to do so from the training data provided.

\subsection{Color Image Forensics}

\begin{figure}[t]
\begin{center}
\includegraphics[width=0.75\linewidth]{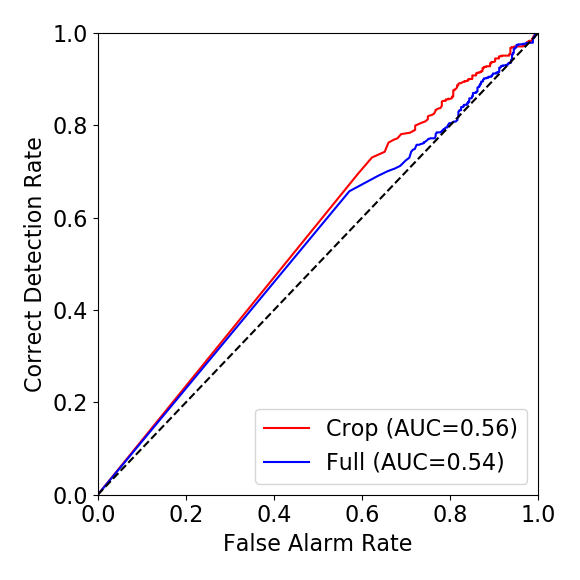}
\end{center}
\caption{ROC curves showing the performance of the color image forensic to the GAN datasets.}
\label{fig:colorROC}
\end{figure}

Unfortunately, the ROC curves from the color image forensic, shown in Fig. \ref{fig:colorROC}, are little better than random.  With AUCs of 0.56 and 0.54 on the two datasets, there isn't much evidence that the classifier learned anything useful about the color statistics of GAN- versus camera-generated imagery.  One possible reason for this is that some of the camera imagery in the evaluation sets contain images with the type of focus manipulations that the pre-trained INH network was intended to detect.  As an example, Fig. \ref{fig:saturationFigure} (top right) shows a picture in which the left and top edges appear to be blurred.  Since these images are taken from a set of celebrity face images, it may be that they have been re-touched in ways that INH considers a manipulation.  This could potentially be addressed by re-training the entire network, though that would require much more training data than was used in the present experiments.

\section{Conclusion}

We have described and evaluated the efficacy of two different forensics related to the way in which generator networks from GANs transform feature representations to red, green, and blue pixel intensities.  We demonstrated, in particular, that a relatively simple forensic based on the frequency of over-exposed pixels provides good discrimination between GAN-generated and camera imagery, via experiments with an independently-generated challenge dataset.  Our method does quite well distinguishing fully GAN-generated image from natural images, and that it still provided some discrimination in the more difficult case where GAN-generated faces are spliced into a larger camera image.  A second forensic, based on color image statistics, proved less useful than saturation statistics, but was likely limited by the lack of available training data.

Both of these forensics were proposed based on a thorough analysis of the generator's architecture, specifically how it transforms a multi-channel feature map into a 3-channel color image.  We show that the weights applied to the features are analogous to how color filter arrays integrate over the visible spectrum, but that they use very different weights.  Our saturation forensic was based on the fact that generators incorporate normalization which limits the range of generated intensities, a limitation which is not present in the irradiance of natural scenes.  We have developed these forensics by targeting operations common to multiple generator architectures, including very recent work from the last year.

That said, the pace of GAN-related innovation is quite high, and it's hard to predict how generators will be structured in the years ahead.  We hope that the type of analysis here could encourage similar analyses of future GANs, in order to help address an important problem impacting increasingly large parts of society.

\ifwacvfinal
\section*{Acknowledgement}
This material is based upon work supported by the United States Air Force and the Defense Advanced Research Projects Agency under Contract No. FA8750-16-C-0190. Any opinions, findings and conclusions or recommendations expressed in this material are those of the author(s) and do not necessarily reflect the views of the United States Air Force or the Defense Advanced Research Projects Agency.
\fi

{\small
\bibliographystyle{ieee}
\bibliography{forensics}

\begin{thebibliography}{10}\itemsep=-1pt

\bibitem{CanChenCVPR2018}
C.~Chen, S.~McCloskey, and J.~Yu.
\newblock Focus manipulation detection via photometric histogram analysis.
\newblock In {\em The IEEE/CVF Conference on Computer Vision and Pattern
  Recognition (CVPR)}, 2018.

\bibitem{CanonSpectral}
dpreview.com.
\newblock Canon spectral responses.
\newblock https://www.dpreview.com/forums/post/32277783.

\bibitem{GAN_NIPS}
I.~Goodfellow, J.~Pouget-Abadie, M.~Mirza, B.~Xu, D.~Warde-Farley, S.~Ozair,
  A.~Courville, and Y.~Bengio.
\newblock Generative adversarial nets.
\newblock In Z.~Ghahramani, M.~Welling, C.~Cortes, N.~D. Lawrence, and K.~Q.
  Weinberger, editors, {\em Advances in Neural Information Processing Systems
  27}, pages 2672--2680. Curran Associates, Inc., 2014.

\bibitem{guera}
D.~Guera and E.~J. Delp.
\newblock Deepfake video detection using recurrent neural networks.
\newblock In {\em IEEE International Conference on Advanced Video and
  Signal-based Surveillance (to appear)}, 2018.

\bibitem{hsu}
C.-C. {Hsu}, C.-Y. {Lee}, and Y.-X. {Zhuang}.
\newblock {Learning to Detect Fake Face Images in the Wild}.
\newblock {\em ArXiv e-prints}, Sept. 2018.

\bibitem{HuntBook}
R.~Hunt and M.~Pointer.
\newblock {\em Measuring Colour (Fourth Edition)}.
\newblock Wiley, 2011.

\bibitem{nVidia}
T.~Karras, T.~Aila, S.~Laine, and J.~Lehtinen.
\newblock Progressive growing of {GAN}s for improved quality, stability, and
  variation.
\newblock In {\em International Conference on Learning Representations}, 2018.

\bibitem{MNIST}
Y.~Lecun, L.~Bottou, Y.~Bengio, and P.~Haffner.
\newblock Gradient-based learning applied to document recognition.
\newblock In {\em Proceedings of the IEEE}, pages 2278--2324, 1998.

\bibitem{albanyEyes}
Y.~Li, M.-C. Chang, and S.~Lyu.
\newblock In ictu oculi: Exposing ai created fake videos by detecting eye
  blinking.
\newblock In {\em International Workshop on Information Forensics and
  Security}, 2018.

\bibitem{marra}
F.~Marra, D.~Gragnaniello, D.~Cozzolino, and L.~Verdoliva.
\newblock Detection of gan-generated fake images over social networks.
\newblock In {\em 2018 IEEE Conference on Multimedia Information Processing and
  Retrieval (MIPR)}, 2018.

\bibitem{checkerboard}
A.~Odena, V.~Dumoulin, and C.~Olah.
\newblock Deconvolution and checkerboard artifacts.
\newblock {\em Distill}, 2016.

\bibitem{MFC18}
N.~I. of~Standards and Technology.
\newblock Media forensics challenge.
\newblock https://www.nist.gov/itl/iad/mig/media-forensics-challenge-2018.

\bibitem{HDRbook}
E.~Reinhard, G.~Ward, S.~Pattanaik, and P.~Debevec.
\newblock {\em High Dynamic Range Imaging: Acquisition, Display, and
  Image-Based Lighting (The Morgan Kaufmann Series in Computer Graphics)}.
\newblock Morgan Kaufmann Publishers Inc., San Francisco, CA, USA, 2005.

\bibitem{NIPS2016_6125}
T.~Salimans, I.~Goodfellow, W.~Zaremba, V.~Cheung, A.~Radford, X.~Chen, and
  X.~Chen.
\newblock Improved techniques for training gans.
\newblock In D.~D. Lee, M.~Sugiyama, U.~V. Luxburg, I.~Guyon, and R.~Garnett,
  editors, {\em Advances in Neural Information Processing Systems 29}, pages
  2234--2242. Curran Associates, Inc., 2016.

\bibitem{VGG}
K.~Simonyan and A.~Zisserman.
\newblock Very deep convolutional networks for large-scale image recognition.
\newblock {\em CoRR}, abs/1409.1556, 2014.

\bibitem{wang2018pix2pixHD}
T.-C. Wang, M.-Y. Liu, J.-Y. Zhu, A.~Tao, J.~Kautz, and B.~Catanzaro.
\newblock High-resolution image synthesis and semantic manipulation with
  conditional gans.
\newblock In {\em Proceedings of the IEEE Conference on Computer Vision and
  Pattern Recognition}, 2018.

\bibitem{LSUN}
F.~{Yu}, A.~{Seff}, Y.~{Zhang}, S.~{Song}, T.~{Funkhouser}, and J.~{Xiao}.
\newblock {LSUN: Construction of a Large-Scale Image Dataset using Deep
  Learning with Humans in the Loop}.
\newblock {\em ArXiv e-prints}, June 2016.

\end{thebibliography}
}

\end{document}